\documentclass[letterpaper, 10 pt, conference]{ieeeconf}

\IEEEoverridecommandlockouts
\usepackage[hidelinks]{hyperref}
\usepackage{cite}
\usepackage{amsmath,amssymb,amsfonts}
\usepackage{algorithmic}
\usepackage{graphicx}
\usepackage{textcomp}
\usepackage{xcolor}
\usepackage{booktabs}
\usepackage{float}

\begin{document}

\title{doPlan: A Variable-Horizon Dataset for Multi-Stage Language-Conditioned Planning in Autonomous Driving}

\author{
Parthib Roy$^{1}$,
Yash Tandon$^{1, 2}$,
Marcus Blennemann$^{1, 2}$,
Giovanni Tapia Lopez$^{1}$,\\
Angel Martinez-Sanchez$^{1}$,
Mohan M. Trivedi$^{2}$,
Ross Greer$^{1}$%
\thanks{$^{1}$Machine Intelligence, Interaction, and Imagination (Mi$^3$) Laboratory, University of California, Merced, Merced, CA, USA.}%
\thanks{$^{2}$Laboratory for Intelligent \& Safe Automobiles (LISA), University of California, San Diego, La Jolla, CA, USA.}%
}

\maketitle

\begin{abstract}
Autonomous vehicles interacting with passengers through natural language must reason beyond immediate commands. Passenger intent may span multiple stages of behavior, depend on future events, refer to surrounding agents or landmarks, and remain relevant as driving conditions evolve. Existing language-enabled driving datasets largely focus on short, localized interactions, leaving these longer-horizon forms of passenger intent comparatively underexplored. We
introduce \textbf{doPlan}, to our knowledge the first publicly available,
human-annotated real-world dataset designed to study passenger language as
persistent task context. Built on nuPlan, doPlan contains 5,154 human-written
passenger instructions spanning 169.1 hours of cumulative instruction-aligned context over 50.9 hours of unique driving, with annotation windows ranging from 30.0 to 508.8\,s. The annotations capture immediate, deferred, event-conditioned, persistent, and multi-stage passenger intent. The dataset, annotation interface, and supporting resources are publicly available at \url{https://github.com/Mi3-Lab/doPlan}. We evaluate four language-conditioned driving models and find that sensitivity to passenger language does not reliably translate into behavior consistent with the requested direction. More broadly, among 2,161 examples with a matched future maneuver, the first associated maneuver occurs a median of 24.6\,s after the evaluation point, and only 9.8\% occur within the models' common 5\,s prediction horizon. These findings highlight the need to connect persistent passenger intent with successive planning decisions. doPlan provides a setting for studying how unresolved goals can be retained, grounded in evolving scenes, and tracked across multiple stages, including how a planner determines when a future goal becomes relevant to the current plan.


\end{abstract}

\section{Introduction}
As robots expand from performing structured tasks, such as package sorting and repetitive component assembly, to operating in everyday human environments (e.g., assisting with household chores and navigating public roads), they will increasingly need to understand not only what a person wants them to do now, but what that person is trying to accomplish beyond the next action. Human instructions often communicate such broader intent through contextual or temporally extended language, in which some parts may be immediately actionable while others depend on how the surrounding situation unfolds. For example, a person might ask a household robot to “bring me the mug after the kettle boils.” Following even this simple request requires the robot to ground the referenced objects, retain a pending goal, recognize when the relevant condition has been satisfied, and act at the appropriate time. Natural language provides a flexible way to communicate such intent because it can express goals, constraints, references, preferences, and sequences of behavior without relying on a predefined command vocabulary. In prior work, language has been used to specify high-level robotic goals that unfold over multiple actions and to connect perception, reasoning, and physical action~\cite{ichter2022saycan,innermonologue,driess2023palme,zitkovich2023rt2}. However, as instructions persist while the environment changes, the challenge extends beyond interpreting what was said: a system must preserve unresolved intent, ground that intent in new observations, and determine when each part of the instruction should influence the current action.

\begin{figure}[t]
    \centering
    \includegraphics[
        width=\columnwidth,
        height=0.28\textheight,
        keepaspectratio
    ]{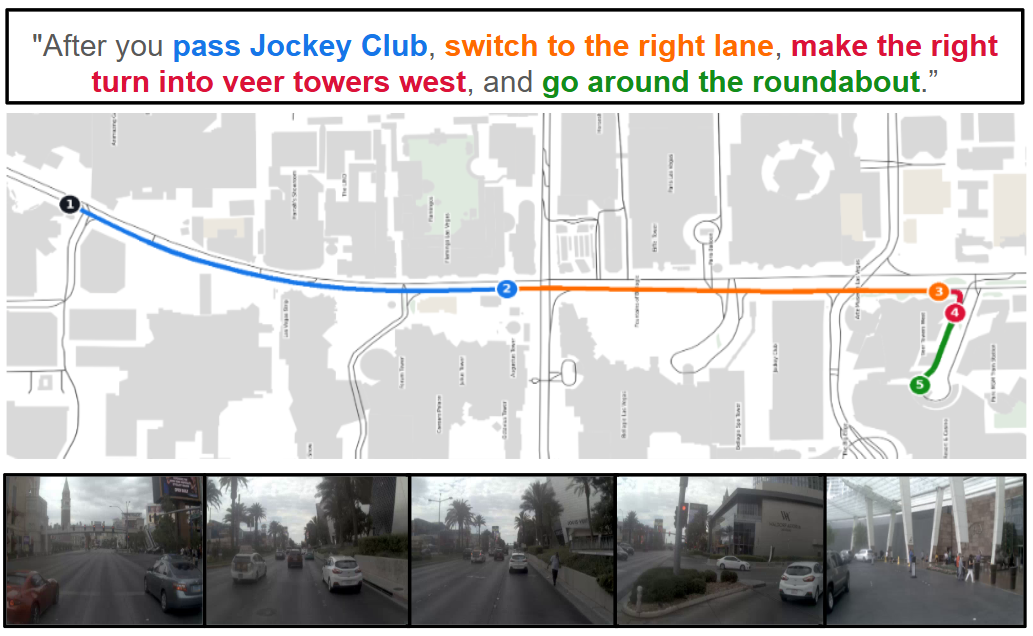}
    \caption{Representative doPlan annotation illustrating temporally extended, multi-stage passenger intent. A single human-written instruction is aligned with successive route segments, with front-camera frames showing the corresponding driving context.}
    \label{fig:doplan_overview}
\end{figure}

In autonomous driving, this problem becomes especially tangible because
passenger instructions can span different temporal horizons. Some requests
describe a near-term maneuver, such as ``turn left at this intersection,''
while others span multiple stages of the drive. Consider a passenger who
says, ``stay in this lane until we pass the construction, then get over for
the next exit.'' The vehicle must maintain the current lane, recognize when
the construction has been passed, retain the later lane-change goal, account
for surrounding traffic, and transition when appropriate. As autonomous
vehicles take on more low-level driving decisions, passenger requests may
increasingly operate at a higher level, expressing goals, preferences, and
contingencies rather than individual maneuvers. In this setting, passenger
language becomes persistent task context whose relevance to immediate motion
changes as the scene evolves.

These requirements arise within a broader literature on language-enabled autonomous driving. Prior work has used language to explain model behavior~\cite{kim2018bddx}, ground human advice and passenger commands~\cite{kim2019had,deruyttere2019talk2car}, answer questions about driving scenes~\cite{DRAMA,driveLM,lingoQA, shriram2025towards}, and condition predicted trajectories or driving actions~\cite{deruyttere2022talk2car,coVLA}. Collectively, these efforts establish language as a useful means of understanding driving scenes and specifying or influencing vehicle behavior. However, they largely study localized scenes or short-horizon interactions. The temporal setting considered here is different: passenger intent may remain relevant while the corresponding behavior is not yet actionable, leaving instructions that persist across multiple stages of an evolving drive comparatively underexplored.

Closest to this setting, doScenes introduced free-form human passenger instructions for 1,000 real-world nuScenes clips~\cite{roy2025doscenes}. Using the ``taxi test,'' annotators describe what they would have told a driver to produce the observed behavior, grounding language in intended vehicle behavior rather than only in scene description. However, doScenes associates each instruction with a fixed 12\,s driving clip, leaving passenger intent that unfolds across longer and multi-stage driving contexts largely outside its scope. Subsequent work showed that these instructions can influence VLA trajectory predictions when used as conditioning input~\cite{doplanapplication}. This leaves a natural temporal question: what changes when passenger intent persists beyond a single short driving scene?

We introduce \textbf{doPlan}, a human-annotated dataset built on the nuPlan driving dataset~\cite{caesar2021nuplan} for studying free-form passenger instructions over extended and evolving driving contexts. doPlan contains 5,154 human-written passenger instructions spanning 169.1 hours of cumulative instruction-aligned context across 50.9 hours of unique annotated driving, anchored to variable-duration windows ranging from 30.0 to 508.8~s. Following the taxi-test principle, annotators write what they would have told a driver to produce the behavior observed within each sampled segment. Because these windows are long enough for behavior to unfold in stages, the resulting instructions include ordered maneuvers, deferred actions, and behaviors that persist until a condition is satisfied.

We evaluate OpenEMMA~\cite{xing2024openemma}, AutoVLA~\cite{autovla}, Alpamayo 1.5~\cite{nvidia2025alpamayo}, and Alpamayo 2.0~\cite{nvidia2026alpamayo2} under matched language conditions to characterize how current language-conditioned driving models use passenger instructions. Two gaps emerge. First, language sensitivity does not reliably translate into instruction following: stronger conditioning can alter predicted trajectories without improving directional compliance. Second, the behaviors associated with passenger instructions often lie well beyond current short-horizon prediction windows. Together, these results motivate treating passenger language as persistent task context rather than as a one-shot perturbation to the next trajectory.


\section{Related Work}

Large-scale autonomous-driving datasets such as KITTI~\cite{kitti}, the Waymo
Open Dataset~\cite{sun2020scalability}, and nuScenes~\cite{caesar2020nuscenes}
have established benchmarks for perception, forecasting, and multimodal scene
understanding, while nuPlan extends this setting toward motion planning over
continuous real-world driving sequences~\cite{caesar2021nuplan}. doPlan builds
on nuPlan by adding free-form passenger instructions aligned with
variable-duration driving contexts.

Natural language has increasingly moved from describing and reasoning about
driving scenes toward directly influencing vehicle behavior. Earlier work
studied explanations, human advice, scene understanding, and grounded passenger
commands~\cite{kim2018bddx,kim2019had,deruyttere2019talk2car,driveLM,lingoQA},
while more recent approaches use language for instruction-conditioned planning,
closed-loop driving, language--action alignment, and vision--language--action
modeling~\cite{lmdrive,lampilot,simlingo,coVLA,zuo2026vegalearningdrivenatural}.
For example, the Honda Research Institute Advice Dataset (HAD)~\cite{kim2019had} contains 30 hours of driving video data with associated natural language advices which give a goal for or otherwise inform the driving behavior. These works establish the importance of language-conditioned driving, but
generally study instructions within relatively bounded scenes or planning
episodes.

Longer-term temporal structure has been explored more directly in navigation
and robotics. Talk2Nav studies long-range route instructions whose components
become relevant as navigation progresses~\cite{talk2nav}, while Lang2LTL
grounds ordered and temporally constrained natural-language tasks into formal
specifications~\cite{lang2ltl}. Within autonomous driving, doScenes is closest
to our passenger-oriented setting, collecting free-form human instructions for
fixed 12-second nuScenes clips using the taxi-test protocol~\cite{roy2025doscenes}.
doPlan follows the same annotation principle but extends it to continuous
nuPlan contexts spanning 30.0--508.8~s, enabling the study of passenger intent
that may span multiple stages, remain pending until future events occur, or
persist as driving conditions evolve.

\section{Methods}


\subsection{Dataset Source and Annotation Interface}

doPlan is constructed exclusively from the official nuPlan training split,
which provides continuous real-world driving logs with synchronized camera and
vehicle trajectory data. Camera streams are preprocessed into synchronized video
segments while preserving nuPlan's native temporal indexing, maintaining
alignment between each annotation and the corresponding portion of the
recorded drive. 

Annotations are collected through a custom interface that presents eight
synchronized camera views and a heading-aligned map of the surrounding road
network. The camera views provide front, side, and rear context, while the map
and compass provide additional information about road geometry, upcoming turns,
cardinal direction, and nearby landmarks. This context helps annotators formulate
passenger instructions that reflect the broader spatial and driving situation
shown in the segment.




\subsection{Segment Sampling Strategy}

To generate annotation candidates, doPlan samples variable-length intervals
from continuous nuPlan driving sequences. For a sequence of duration $T$, we
first sample a segment duration $L \sim \mathrm{Uniform}(30,T)$ and then sample
its start time $S \sim \mathrm{Uniform}(0,T-L)$. This produces contexts ranging
from relatively short maneuvers to substantially longer driving episodes. We
sample directly from continuous trajectories without preselecting maneuver or
scenario categories, and sampled intervals may overlap by design, allowing partially
shared driving contexts to receive independent passenger instructions. The
30\,s minimum provides sufficient temporal context for extended instruction
generation while preserving broad scenario coverage.









\subsection{Instruction Annotation Protocol}
\label{sec:annotation_protocol}

doPlan uses a passenger-oriented annotation protocol based on the ``taxi
test.'' As annotators review each driving segment, they formulate an
instruction they could plausibly have given to a driver to produce the
observed driving behavior. They are asked:

\begin{quote}
\emph{``Imagine you are a passenger in this taxi. What instruction would you
give the driver to produce the behavior you see in the video?''}
\end{quote}

This framing is intended to elicit instructions that a passenger could
reasonably give to a driver, rather than descriptions of the scene or direct
summaries of the recorded trajectory. Annotations are collected without a
predefined template or command vocabulary and may take several forms:
(i) a single-clause instruction (e.g., ``turn left at the next intersection''),
(ii) a multi-clause or multi-sentence instruction when the observed behavior
spans multiple stages (e.g., ``stay in this lane until we pass the construction,
then move left for the next exit''), or (iii) a no-instruction submission when
the annotator does not identify a plausible passenger directive for the observed
behavior nor a need to provide instruction based on the scene simplicity. No-instruction submissions are retained as valid annotation outcomes
for accounting purposes but are excluded from the passenger-instruction set.

The sampled temporal window defines the driving context shown to the annotator; it does not specify when a passenger would have given the instruction during the drive. In particular, the first frame of the window should not be interpreted as the moment the instruction was spoken. The annotation protocol allows instructions to refer to information that a passenger could reasonably know from the route or surrounding environment, including landmarks, upcoming turns, and other contextual information beyond what is visible in a single frame.

Nineteen annotators, all licensed drivers, contributed to the dataset. They did not
have access to previous annotations associated with a segment, encouraging independent
linguistic formulations. Because multiple natural-language instructions may be
behaviorally consistent with the same observed trajectory, doPlan does not assume a
unique ground-truth instruction.

Each annotation stores the temporal window, free-form instruction, referential label, and anonymized annotator identifier. We remove exact duplicates and records with invalid or missing fields while retaining distinct instructions associated with the same temporal window.

\subsection{Referential Labels}


Following doScenes~\cite{roy2025doscenes}, each instruction is assigned a coarse referential label describing the type of environmental reference it contains. These labels provide categorical rather than fine-grained word-object or bounding-box-level grounding. The label definitions and resulting dataset distribution are summarized in Table~\ref{tab:referential_distribution}. 

\section{Dataset Analysis}

The annotation process produced 6,549 submissions, of
which 5,154 contain non-empty human-written passenger instructions and 1,395
contain no instruction text. The passenger instructions span 169.1 hours
of cumulative instruction-aligned context and 50.9 hours of unique annotated
driving. These annotations are drawn from 988 nuPlan source clips.

\subsection{Scale and Temporal Coverage}

doPlan extends the short-term interaction setting introduced by
doScenes~\cite{roy2025doscenes} to substantially longer and variable driving
contexts. Table~\ref{tab:doscenes_doplan} summarizes the main differences in
scale and temporal coverage between the two datasets.

\begin{table}[!ht]
\centering
\caption{Comparison of doScenes and doPlan.}
\label{tab:doscenes_doplan}
\begin{tabular}{lcc}
\toprule
\textbf{Property} & \textbf{doScenes} & \textbf{doPlan} \\
\midrule
Passenger instructions & 2,450 & 5,154 \\
Window duration & 12~s & 30.0--508.8~s \\
Cumulative instruction context & $\sim$8.17~h & 169.1~h \\
Unique annotated context & $\sim$2.4~h & 50.9~h \\
Temporal horizon & Fixed & Variable \\
\bottomrule
\end{tabular}
\end{table}

\begin{figure}[t]
    \centering
    \includegraphics[width=\columnwidth]{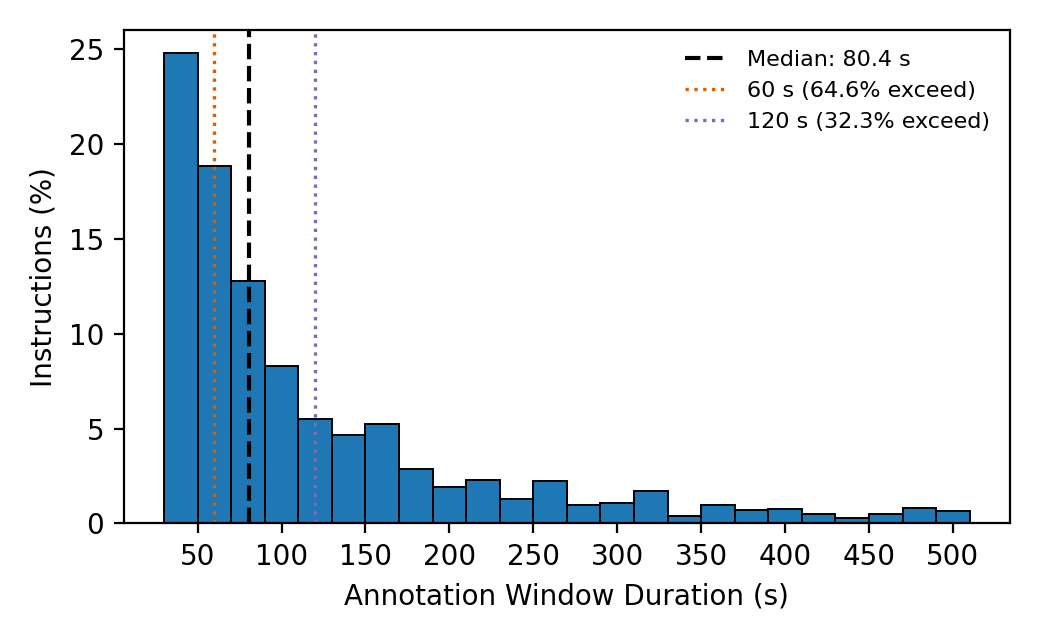}
    \caption{Distribution of instruction-bearing window durations ($n=5{,}154$).
    The median is 80.4\,s; 64.6\% of windows exceed 60\,s and 32.3\% exceed 120\,s.}
    \label{fig:window-duration}
\end{figure}

Within doPlan, instruction-bearing windows span a broad range of temporal horizons (Fig.~\ref{fig:window-duration}).

\subsection{Temporal Structure and Instruction Complexity}

To examine how instruction structure varies with temporal context, we divide the
5,154 instruction-bearing annotations into five approximately equal groups
based on annotation-window duration. Table~\ref{tab:duration_complexity}
reports the mean number of words per instruction, the percentage containing
multiple sentences, and a simple proxy for multi-step language. For this
analysis, an instruction is counted as multi-step if it contains multiple
sentences or the temporal marker \textit{then}.

All three measures increase with annotation-window duration. Instructions in
the shortest-duration group contain an average of 9.8 words, with 24.2\%
containing multiple sentences and 28.8\% satisfying the multi-step proxy. In
the longest-duration group, these values increase to 27.8 words, 52.3\%, and
62.6\%, respectively. These trends indicate that longer driving contexts are
associated with longer and more structurally complex passenger instructions,
rather than merely providing additional video around otherwise similar
commands.

\begin{table}[t]
\centering
\caption{Instruction structure across annotation-window duration groups in
doPlan.}
\label{tab:duration_complexity}
\begin{tabular}{lccc}
\toprule
\textbf{Annotation Window (s)} &
\textbf{Mean Words} &
\textbf{Multi-Sentence} &
\textbf{Multi-Step} \\
\midrule
30.0--45.2 ($n=1{,}031$)    & 9.8  & 24.2\% & 28.8\% \\
45.2--65.0 ($n=1{,}031$)    & 11.2 & 27.8\% & 35.1\% \\
65.2--97.4 ($n=1{,}031$)    & 13.6 & 33.3\% & 48.1\% \\
97.4--169.8 ($n=1{,}031$)   & 17.7 & 46.4\% & 53.2\% \\
169.8--508.8 ($n=1{,}030$)  & 27.8 & 52.3\% & 62.6\% \\
\bottomrule
\end{tabular}
\end{table}

\subsection{Referential and Linguistic Structure}

\begin{table}[t]
\centering
\caption{Referential labels and their distribution in doPlan.}
\label{tab:referential_distribution}
\scriptsize
\setlength{\tabcolsep}{3pt}
\begin{tabular}{@{}p{2.15cm}p{3.15cm}rr@{}}
\hline
\textbf{Label} & \textbf{Description} & \textbf{Count} & \textbf{\%} \\
\hline
Non-Referential &
No specific scene reference. &
782 & 15.17 \\

Static &
Fixed scene element. &
2,749 & 53.34 \\

Dynamic &
Moving agent. &
629 & 12.20 \\

Both &
Static and dynamic elements. &
954 & 18.51 \\

Ambiguous &
Reference type is unclear. &
39 & 0.76 \\

Other/Malformed &
Other or malformed label. &
1 & 0.02 \\
\hline
\textbf{Total} & & \textbf{5,154} & \textbf{100.00} \\
\hline
\end{tabular}
\end{table}

The instructions also vary in how they refer to the surrounding
driving environment. Table~\ref{tab:referential_distribution}
summarizes the distribution of annotator-assigned referential labels.
Static references are the most common, followed by instructions
containing both static and dynamic references. Overall, 84.1\% of
instructions are labeled as static, dynamic, or both, indicating that
a large fraction of doPlan instructions depend on surrounding agents,
landmarks, or other elements of the driving environment. Though this dataset is language-only, we note that referential language can also be supplemented by multimodal observation of the instruction-giver, creating a link between the user may be looking at or pointing to and the language of their instruction \cite{bossen2025can}. 

\subsection{Temporal Context Validation}

A central design choice in doPlan is the use of substantially longer
driving contexts than those provided by prior instruction-conditioned driving datasets.
An important question is whether these extended windows contain
additional information related to the passenger instruction, rather
than simply adding more video around the same short-term behavior.
We therefore use a learned video--instruction matcher built on CLIP to test
whether additional temporal context contains instruction-relevant information.
The matcher extracts visual features from front-camera frames and text features
from each candidate instruction, then learns to assign higher similarity to
matching video--instruction pairs than to mismatched pairs.

For each driving sequence, the matcher ranks 100 human-written
instructions: the associated doPlan instruction and 99 instructions
drawn from different nuPlan source logs. Instructions from the same
source log are excluded from the negative set to avoid treating another
potentially valid annotation of overlapping driving context as
incorrect.


Table~\ref{tab:retrieval_context} shows that retrieval improves with
additional driving context. R@1 increases from 13.9\% using the first
5\,s to 23.1\% with the full window, while R@5 increases from 38.2\%
to 54.2\% and R@10 from 52.2\% to 68.6\%.

\begin{table}[t]
\centering
\caption{Retrieval of the correct passenger instruction from 100
candidates as more of the driving sequence is observed.}
\label{tab:retrieval_context}
\small
\begin{tabular}{lccc}
\toprule
\textbf{Video observed} & \textbf{R@1} & \textbf{R@5} & \textbf{R@10} \\
\midrule
5 s         & 13.9\% & 38.2\% & 52.2\% \\
15 s        & 14.9\% & 40.4\% & 55.5\% \\
30 s        & 16.5\% & 43.3\% & 57.5\% \\
60 s        & 17.8\% & 44.7\% & 59.9\% \\
Full window & \textbf{23.1\%} & \textbf{54.2\%} & \textbf{68.6\%} \\
\bottomrule
\end{tabular}
\end{table}

The improvement is not explained by simple dataset properties.
Duration-only, text-only, and frozen CLIP~\cite{radford2021learningtransferablevisualmodelsclip}
baselines obtain 2.7\%, 1.3\%, and 3.5\% R@1, respectively, compared
with 23.1\% for the learned matcher. Together, these results provide evidence that doPlan's
extended annotation windows contain instruction-relevant information
beyond what is available in the initial few seconds of driving. We use
this experiment as a validation of the dataset's temporal design.

\section{Model Evaluation and Diagnostic Analysis}
\label{sec:benchmark_evaluation}

As a natural starting point, we examine whether contemporary
language-conditioned driving models meaningfully use the passenger
instructions in doPlan when predicting vehicle trajectories. OpenEMMA,
AutoVLA, Alpamayo~1.5, and Alpamayo~2.0 were evaluated zero-shot using their
native sensor configurations. For each evaluated instruction, we evaluate the models at the
earliest time point within the annotated driving segment for which the required
sensor history and future trajectory are available. This evaluation time point
coincides with the start of the annotation window in 85.0\% of cases; otherwise,
it occurs 0.2--4.6\,s later, with a median offset of 2.8\,s. For a given model
and example, the same sensor observations at this time point are used across
all language conditions, such that only the language input changes. This controlled setup allows us to ask whether changes in predicted behavior can be attributed to the passenger language itself. This
evaluation time point serves as the common reference for trajectory prediction
and subsequent temporal analyses and should not be interpreted as the time at
which a passenger would have spoken the instruction.

From the full dataset of 5,154 instructions, we evaluate a subset of
4,334 examples.\footnote{We use a subset due to limited shared compute.
Alpamayo~2.0 was run on two NVIDIA H200 GPUs and required four
generations per example.}
Of these, 4,315 are used for direct cross-model comparison; the remaining
19 were excluded because AutoVLA did not produce valid outputs in the
required action format.

For each model, we evaluate four conditions designed to
distinguish instruction-specific behavior from generic sensitivity to language:
no instruction; the corresponding human-written passenger instruction; an
unrelated human-written instruction drawn from a different clip (e.g., a
passenger instruction associated with another driving scene); and a rule-based
counterfactual intended to request different behavior in the same scene
(e.g., ``change to the right lane'' becomes ``change to the left lane'').
All predicted trajectories are standardized to a common egocentric
representation at ten 0.5\,s intervals over a 5\,s horizon, the longest horizon
shared by all four models.

\begin{table}[t]
    \centering
\caption{Trajectory alignment across language conditions on 4,315
instructions. ADE (average displacement error) is measured
over a common 5\,s horizon.}
    \label{tab:ade_results}
    \scriptsize
    \setlength{\tabcolsep}{4pt}

    \begin{tabular}{lcccc}
        \toprule
        \textbf{Model} &
        \multicolumn{4}{c}{\textbf{ADE $\downarrow$ (m)}} \\
        \cmidrule(lr){2-5}
        &
        \textbf{No Instr.} &
        \textbf{Correct} &
        \textbf{Unrelated} &
        \textbf{Counterfactual} \\
        \midrule
        OpenEMMA
            & 3.342
            & \textbf{3.307}
            & 3.359
            & 3.340 \\
        AutoVLA
            & \textbf{2.663}
            & 2.784
            & 2.846
            & 2.872 \\
        Alpamayo~1.5 ($w=1.0$)
            & \textbf{1.979}
            & 2.100
            & 2.106
            & 2.081 \\
        Alpamayo~2.0 ($w=3.0$)
            & \textbf{1.451}
            & 1.834
            & 1.816
            & 1.858 \\
        \bottomrule
    \end{tabular}


\end{table}

Table~\ref{tab:ade_results} shows that providing the corresponding passenger instruction does not consistently improve trajectory alignment over the no-instruction baseline. OpenEMMA improves slightly by 1.1\%, while AutoVLA and Alpamayo~1.5 worsen by 4.5\% and 6.1\%, respectively. Alpamayo~2.0 shows the largest degradation, with ADE increasing by 26.4\% from 1.451\,m to 1.834\,m. This is consistent with prior evaluation on doScenes, where passenger instructions influenced OpenEMMA trajectories without consistently improving trajectory alignment~\cite{doplanapplication}.

More importantly, the correct passenger instruction produces ADE values similar to those obtained with unrelated and counterfactual instructions across all four models, suggesting that the models do not reliably distinguish the intended instruction from semantically mismatched language. One possible explanation is that the language signal is simply too weak to meaningfully influence the predicted trajectory.

\subsection{Navigation Guidance and Language Sensitivity}

We examine this possibility using Alpamayo~2.0, whose navigation-guidance strength can be varied at inference time. This allows us to test whether increasing the strength of language conditioning leads to greater separation between trajectories produced under different instructions.

We evaluate guidance weights from $w=1.0$ to $w=4.0$ over the same 4,315 evaluation instructions, with $w=3.0$ corresponding to Alpamayo~2.0's released default setting. Across guidance settings, the model reasoning output and diffusion noise are held fixed, leaving the guidance weight as the only varying factor. Because ADE measures agreement with the recorded human trajectory, we introduce a separate measure of how strongly replacing the original instruction with its counterfactual changes the model's prediction. We define \emph{trajectory separation} as

\begin{equation}
S =
\frac{1}{10}
\sum_{k=1}^{10}
\left\|
\hat{p}^{\,\mathrm{orig}}(t_k)
-
\hat{p}^{\,\mathrm{cf}}(t_k)
\right\|_2 ,
\end{equation}

where $t_k \in \{0.5,1.0,\ldots,5.0\}$\,s and $\hat{p}(t_k)$ denotes the
predicted ego position at time $t_k$. Unlike ADE, trajectory separation
does not use the recorded human trajectory. A larger value means that
replacing the original instruction with its counterfactual produces a larger
change in the predicted trajectory; it does not indicate that either
trajectory is more correct.

\begin{table}[!ht]
    \centering
    \caption{Effect of Alpamayo~2.0 navigation-guidance weight on trajectory separation.}
    \label{tab:alpamayo_guidance}
    \scriptsize
    \setlength{\tabcolsep}{2pt}
    \begin{tabular*}{\columnwidth}{@{\extracolsep{\fill}}lccccccc@{}}
        \toprule
        \textbf{$w$}
        & 1.0 & 1.5 & 2.0 & 2.5 & 3.0 & 3.5 & 4.0 \\
        \midrule
        \textbf{$S$ (m)}
        & 0.219 & 0.297 & 0.375 & 0.455 & 0.533 & 0.607 & 0.679 \\
        \bottomrule
    \end{tabular*}
\end{table}

Trajectory separation increases monotonically from 0.219 to 0.679\,m across the evaluated guidance range. At the released default setting of $w=3.0$, separation reaches 0.533\,m. These results show that stronger navigation guidance consistently increases the model's sensitivity to changes in passenger language. However, greater trajectory sensitivity does not establish that the model responds in the direction requested by the passenger. We therefore next evaluate whether this increased sensitivity corresponds to directional compliance.

\begin{figure}[t]
    \centering
    \includegraphics[
        width=\columnwidth,
        height=0.30\textheight,
        keepaspectratio
    ]{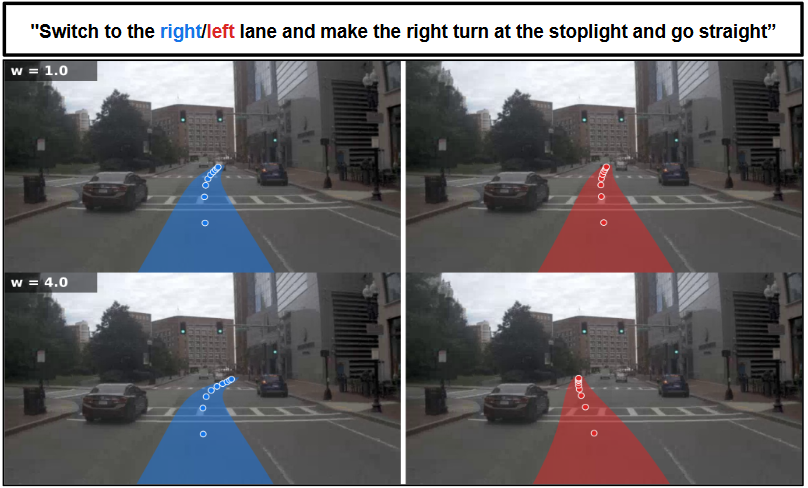}
\caption{Increasing navigation guidance makes Alpamayo~2.0 more responsive
to passenger language. For the same driving scene, replacing the original
``right lane'' instruction (blue) with its left--right counterfactual,
``left lane'' (red), yields a trajectory separation of $S=0.95$\,m at
$w=1.0$ and $S=3.84$\,m at $w=4.0$. At higher guidance, the counterfactual
trajectory shifts toward the left lane while the original trajectory bends
more sharply to the right.}
    \label{fig:trajectory_separation}
\end{figure}

\subsection{Directional Instruction Compliance}

We find that increasing navigation guidance produces larger changes in the
predicted trajectory, but this does not establish that those changes agree with
the passenger's requested direction. We examine this using a subset of 851
instructions containing a single left- or right-turn request and no additional
detected actions (475 left, 376 right). For 612 of these instructions, we
identify a corresponding turn in the requested direction from the recorded
nuPlan ego trajectory.

We then consider when these matched turns occur relative to the models' predictions and find that only 4.4\% begin within the common 5\,s prediction horizon, revealing a substantial temporal mismatch between the requested maneuver and the trajectory available for evaluation. For most directional instructions, the corresponding turn begins after the predicted trajectory ends. An immediate lateral response is therefore not expected in many cases, and its absence cannot by itself be interpreted as failure to follow the passenger instruction.

We therefore evaluate directional compliance only for the matched turns that
begin within the 5\,s horizon. We measure whether the predicted 5\,s endpoint
lies at least 1\,m toward the requested side, and additionally test a
left--right counterfactual in which the requested direction is reversed while
holding the observation and sampling seed fixed. We measure counterfactual
directional consistency by requiring both the original and reversed
instructions to produce motion toward their respective requested sides.

Even within this temporally relevant subset, stronger language conditioning
provides little evidence of reliable directional following. At the default
guidance setting ($w=3.0$), conditioning on the passenger instruction increases
requested-side response by only 1.3 percentage points over the no-instruction
condition (49.3\% vs.\ 48.0\%). No corresponding increase is observed at
$w=1.0$.

As a complementary check independent of detected turn timing, we identify
129 instructions for which the recorded human trajectory moves at least
1\,m toward the requested side within 5\,s. We evaluate three sampled
predictions per instruction, yielding 387 predictions at each guidance weight.
Counterfactual directional consistency is not observed in any prediction.
Because the mechanically generated counterfactual direction may be infeasible
in some scenes, we treat this as a diagnostic rather than a complete measure
of executable compliance.

These results expose two aspects of the same temporal problem: most requested
turns lie beyond the horizon of the current trajectory prediction, and even
when the requested turn is imminent, stronger language conditioning does not
reliably produce motion in the requested direction. We therefore next examine
how model response changes with temporal distance to the associated maneuver.

\subsection{Temporal Alignment of Future Maneuvers}
\label{sec:temporal_grounding}

The preceding experiments test whether models respond to passenger language,
but not how that response varies with temporal distance to the associated
future maneuver. Our evaluation uses a common 5\,s prediction horizon, while a maneuver associated with an instruction may occur much later relative to the standardized evaluation point.

To characterize this gap, we align 4,312 evaluated instruction annotations with recoverable
annotation timing to their original nuPlan drives and identify when the
corresponding maneuver begins using the recorded ego-pose logs and map data.
We reconstruct the recorded ego trajectory and use trajectory- and map-based
detectors for left and right turns, sustained stops, and straight intersection
traversals. For example, an instruction such as ``turn left'' is matched to
the first detected left-turn onset after the evaluation time point. For each
instruction, we consider the earliest action that can be mapped to one of these
supported event types. Of these annotations, 2,161 have valid nuPlan log alignment
and a detected onset for that action. Measured from the evaluation time point
defined above, these detected events occur a median of 24.6\,s later, and only
9.8\% fall within the models' common 5\,s prediction horizon. Because the
evaluation time point is tied to the sampled annotation window, the exact
offset depends in part on window construction; however, restricting the
analysis to shorter annotation windows yields the same qualitative result:
the associated maneuver typically occurs well beyond the 5\,s prediction
horizon.



This temporal gap raises a more specific question: when a passenger names
a future turn, does the requested direction affect the current plan only
as that turn approaches, or does the model respond even when the maneuver
is still far away?

Of these annotations, 1,483 correspond to left- or right-turn maneuvers.
For each annotation, we compare the trajectory generated by the original
instruction with the trajectory generated after reversing the named direction.
We define directional response as

\begin{equation}
    R_i = d_i \left(
    \theta_i^{\mathrm{orig}} - \theta_i^{\mathrm{rev}}
    \right),
\end{equation}

where $\theta_i^{\mathrm{orig}}$ and $\theta_i^{\mathrm{rev}}$ are the
predicted headings under the original and reversed instructions, respectively,
and $d_i=+1$ for left turns and $d_i=-1$ for right turns. Thus, $R_i>0$
indicates that the original instruction shifts the prediction toward the
requested direction. We group $R_i$ by time to the matched recorded turn.


\begin{table}[!htbp]
    \centering
    \caption{Mean directional response $R$ versus time to the recorded turn.
Positive values indicate a shift toward the requested direction.}
    \label{tab:temporal_directional_response}
    \scriptsize
    \setlength{\tabcolsep}{4pt}
    \begin{tabular}{lrrrr}
        \toprule
        \textbf{Time to turn} &
        \textbf{OpenEMMA} &
        \textbf{Alp.~1.5} &
        \textbf{AutoVLA} &
        \textbf{Alp.~2.0} \\
        \midrule
        $\leq 5$\,s & -0.001 & -0.022 & \textbf{+0.818} & +0.006 \\
        5--10\,s    & +0.003 & +0.037 & \textbf{+0.546} & -0.080 \\
        10--30\,s   & -0.004 & -0.001 & \textbf{+0.282} & +0.051 \\
        30--60\,s   & +0.000 & +0.045 & \textbf{+0.267} & +0.025 \\
        $>60$\,s    & +0.003 & -0.008 & \textbf{+0.199} & +0.057 \\
        \bottomrule
    \end{tabular}
\end{table}

AutoVLA shows the clearest dependence of directional response on temporal
distance to the matched turn. Its current trajectory responds strongly to the
named direction when the turn is imminent, but the effect remains positive even
when the corresponding human turn is more than 60\,s away. OpenEMMA and Alpamayo~1.5 show little consistent directional response,
while Alpamayo~2.0 exhibits a smaller effect.

These results separate two requirements of instruction following:
understanding \emph{what} the passenger wants and understanding \emph{when}
that request should influence the active motion plan. An early directional
response is not necessarily incorrect, since road curvature, lane positioning,
and preparation for an upcoming maneuver may legitimately influence the
trajectory before the turn begins. We therefore treat this experiment as a
diagnostic of temporal grounding rather than proof of premature execution:
a planner may encode the correct future intent without sharply separating
that intent from the behavior that is appropriate now.

\section{Limitations}

The goal of doPlan is to support research on a range of problems in
language-conditioned autonomous driving, including multi-stage instruction
following, deferred and event-conditioned actions, persistent goals,
referential and temporal grounding, instruction generation and composition,
and planning as driving conditions evolve. The dataset provides free-form
passenger instructions aligned with extended driving contexts, but several
limitations should be considered when interpreting the annotations and
benchmark results.

While doPlan uses a common annotation protocol and collects instructions from
licensed drivers, we do not independently revalidate every retained
instruction through a second round of manual review. We remove exact duplicates
and records with malformed or missing fields during cleaning, but some variation
may remain in how annotators describe the same observed behavior, maneuver, or
scene reference. This variation is expected by design because doPlan does not
assume a unique ground-truth instruction for a given observed driving behavior
(Sec.~\ref{sec:annotation_protocol}).

Annotators may also replay the full segment before writing an instruction,
which helps capture richer references, deferred actions, and temporally extended
behavior, but means that referenced information need not be visible at the
evaluation time point. This does not necessarily make the instruction
implausible, since a passenger may rely on route knowledge, familiar landmarks,
or other contextual information beyond the instantaneous scene. However, doPlan
does not currently distinguish such prior knowledge from information visible at
the evaluation time point, nor does it annotate a natural instruction-utterance
or actionability time. Accordingly, temporal offsets reported in our benchmark
should be interpreted relative to the selected evaluation time point rather
than as measured instruction-to-action delays.

\section{Discussion}

The results of doPlan reveal a distinction that is easy to overlook in
language-conditioned driving. Greater sensitivity to passenger language might
appear desirable: if the instruction changes, the model should respond.
Our guidance ablation shows that Alpamayo~2.0 does become progressively more
sensitive as navigation guidance increases. Yet the directional-compliance
results show that larger changes in the predicted trajectory do not reliably
correspond to behavior in the requested direction. This raises a more
fundamental question: what does it mean for a planner to use a passenger
instruction? Responding to a change in language is only one part of the
problem; the planner must also understand what behavior is being requested and
determine how that request should influence the current plan.

The difficulty becomes more pronounced when passenger intent spans different
temporal horizons or multiple stages of a drive\cite{roque2025automated}. Consider ``stay behind the bus
until the intersection, then move left.'' The first part constrains the current
behavior, while the second must remain relevant without yet determining the
immediate motion. Once the intersection is reached, their roles change. This
suggests an \emph{instruction lifecycle} in which different parts of an
instruction may be pending, active, completed, or invalidated as the scene
evolves. Variable-horizon, multi-stage instruction following is therefore not
simply a matter of remembering language for longer. A planner must preserve
unresolved intent while continually determining which parts of that intent are
relevant to the present situation.

This perspective also changes how instruction following should be evaluated.
For a future maneuver, little or no immediate trajectory response may be
appropriate; at the same time, preparation such as lane positioning may begin
well before the maneuver itself. Trajectory error alone therefore captures only
part of the problem, particularly when multiple trajectories may satisfy the
same passenger request. Evaluation across temporal horizons may need to consider
whether requested events occur, whether stages are completed in the correct
order, how model behavior changes as the associated events approach, and
whether the resulting motion remains consistent with safe driving. In this
setting, instruction following concerns not only similarity to a recorded
trajectory, but whether passenger intent is preserved and realized as the drive
unfolds.

This long-horizon setting exposes several research problems that are difficult to study using short, isolated driving clips:

\begin{itemize}
    \item \textbf{Temporal grounding and anticipation:} determine when and how
    a future request should influence the current plan. A maneuver may require
    preparation before it can be executed, making it important to distinguish
    appropriate anticipation from premature action.

    \item \textbf{Persistent intent and progress:} retain unresolved passenger
    goals while tracking which parts of an instruction have become relevant or
    have already been completed as the drive evolves.

    \item \textbf{Event-conditioned execution:} associate requests with the
    future agents, landmarks, or events on which their execution depends, and
    recognize when those conditions have been satisfied.

    \item \textbf{Deferral and revision:} distinguish goals that should affect
    the trajectory now from those that should be deferred, while revising the
    active plan when changing conditions make the original action infeasible.
\end{itemize}

These problems shift the role of passenger language from a command that maps directly to the next trajectory to persistent task context that must be maintained over time and whose influence on the active plan changes as the drive unfolds. This capability is also relevant to safety-critical cases in which a conventional minimal-risk stop may itself create hazards, requiring the vehicle to interpret human guidance and revise its motion through a complex scene~\cite{tandon2026stoppingfailsrethinkingminimal}.

This broader view is also motivated by how passengers naturally communicate
with human drivers. Passengers typically express goals, references,
preferences, and contingencies rather than precise waypoints or control
commands: ``turn after that truck,'' ``stay here until we pass the
construction,'' or ``pull over near the entrance.'' doPlan intentionally
preserves this free-form interaction rather than reducing passenger language to
a fixed navigation vocabulary. Its purpose is not to prescribe a single task
or evaluation protocol, but to provide a setting in which researchers can
study how passenger intent is retained, grounded, revised, and acted upon
across different stages and temporal horizons. More broadly, doPlan supports a
shift from isolated command response toward sustained reasoning over passenger
intent as the driving environment evolves.

\section{Conclusion}

We introduced doPlan, a human-annotated dataset for studying language-conditioned autonomous driving over extended and evolving driving contexts. Its 5,154 free-form passenger instructions span 169.1 hours of cumulative
instruction-aligned context over 50.9 hours of unique annotated driving and
capture a regime in which passenger intent can span multiple stages, depend
on future events, and remain relevant well beyond the next predicted trajectory. Evaluating four contemporary language-conditioned driving models illustrates why this regime matters. Sensitivity to passenger language is not the same as following it, and the maneuvers that instructions refer to typically fall well outside current prediction horizons. Neither gap is addressed by better trajectory regression alone.

These gaps point to a broader planning problem that doPlan is designed to support. By pairing free-form passenger language with extended, continuous driving contexts, doPlan provides a setting for developing models that retain unresolved goals, reason over multi-stage instructions, ground references as scenes evolve, distinguish immediate from future intent, and evaluate behavior over horizons where the requested action may not yet be relevant. More broadly, doPlan aims to move language-conditioned driving from isolated command response toward sustained reasoning over passenger intent as the world evolves.






\bibliographystyle{IEEEtran}
\bibliography{refs}

\end{document}